\documentclass[10pt,twocolumn,letterpaper]{article}

\usepackage[pagenumbers]{cvpr} 

\usepackage[table,dvipsnames]{xcolor}

\usepackage[accsupp]{axessibility} 
\definecolor{cvprblue}{rgb}{0.21,0.49,0.74}
\definecolor{lightsalmon}{rgb}{1.0, 0.63, 0.48}
\usepackage[pagebackref,breaklinks,colorlinks,citecolor=cvprblue]{hyperref}
\usepackage{arydshln}
\usepackage{bbm}
\usepackage{amsthm}
\usepackage[vlined]{algorithm2e}
\usepackage{algorithmic}
\usepackage{multirow}

\newcommand{\N}{\mathbb{N}}		
\newcommand{\R}{\mathbb{R}}		
\newcommand{\p}{\mathbb{P}}		

\renewcommand{\p}{\mathrm{p}}		
\newcommand{\support}{\mathbb{S}}
\newcommand{\query}{\mathbb{Q}}
\newcommand\bs[1]{\boldsymbol{#1}}
\def\sbmm#1{\ensuremath{\boldsymbol{#1}}}          

\newcommand{\syst}[1]{\left \{ \begin{array}{l} #1 \end{array} \right. \kern-\nulldelimiterspace}	

\newcommand{\argmax}{\text{\normalfont argmax}}

\newtheorem{proposition}{Proposition}
\newtheorem{lemma}{Lemma}
\theoremstyle{definition}

\def\paperID{12454} 
\def\confName{CVPR}
\def\confYear{2024}

\title{Transductive Zero-Shot and Few-Shot CLIP}

\author{%
 S\'egol\`ene Martin$^*$, 
  Yunshi Huang$^\dagger$, Fereshteh Shakeri$^\dagger$, Jean-Christophe Pesquet$^*$, Ismail Ben Ayed$^\dagger$  \\ \\
  $^\star$ \emph{Universit\'e Paris-Saclay, Inria, CentraleSup\'elec, CVN} \\ 
  $^\dagger$ \emph{\'ETS Montr\'eal}
}

\begin{document}
\maketitle
\begin{abstract}
Transductive inference has been widely investigated in few-shot image classification, but completely overlooked in the recent, fast growing literature on adapting vision-langage models like CLIP. 
This paper addresses the transductive zero-shot and few-shot CLIP classification challenge, in which inference is performed jointly across a mini-batch of unlabeled query samples, rather than treating each instance independently. We initially construct informative vision-text probability features, leading to a classification problem on the unit simplex set. Inspired by Expectation-Maximization (EM), our optimization-based classification objective models the data probability distribution for each class using a Dirichlet law. The minimization problem is then tackled with a novel block Majorization-Minimization algorithm, which simultaneously estimates the distribution parameters and class assignments. Extensive numerical experiments on 11 datasets underscore the benefits and efficacy of our batch inference approach.
On zero-shot tasks with test batches of 75 samples, our approach yields near 20$\%$ improvement in ImageNet accuracy over CLIP's zero-shot performance. Additionally, we outperform state-of-the-art methods in the few-shot setting. The code is available at: \url{https://github.com/SegoleneMartin/transductive-CLIP}.
\end{abstract}    
\section{Introduction}
\label{sec:intro}

The emergence of large-scale vision-language models like CLIP \cite{radford2021learning} has marked a significant turning point in representation learning \cite{wang2022medclip,jia2021scaling,li2021supervision}. By integrating both visual and textual modalities, these models have shown remarkable potential in crafting generic and richly informative concepts. Unlike traditional vision models, often constrained by task specificity, the representations gleaned from vision-language models are versatile, setting the stage for a breadth of downstream vision tasks and expanding the horizons of what is achievable in the domain.

Among the vision tasks that can be addressed with vision-language models, zero-shot and few-shot classification 
have particularly attracted attention. Notably, CLIP has demonstrated strong performance in zero-shot classification \cite{radford2021learning}. Several subsequent works have leveraged few-shot data, a few labeled samples in the target downstream task, to further improve CLIP's classification accuracy. Following on from the research on prompt learning in the NLP community, CoOp and CoCoOp \cite{zhou2022conditional,zhou2022learning} fine-tuned the pre-trained CLIP model using learnable textual tokens. Another type of approaches, like CLIP-Adapter \cite{gao2023clip} and TIP-Adapter \cite{zhang2022tip} provided CLIP with a parametric feature transformation, which generates adapted features and combines them with the original CLIP-encoded features. Despite their efficacy on few-shot classification benchmarks, these methods predominantly operate within the so-called \emph{inductive} setting, where inference is conducted independently for each query (i.e., test) sample.

In contrast, in the \emph{transductive} paradigm, one makes joint predictions for a batch of query samples, taking advantage of the query set statistics. The transductive setting for few-shot classification with vision-only models was pioneered in \cite{liu2018learning}, and have since become prominent research subject, triggering an abundant, very recent literature on the subject, e.g., \cite{martin2022towards,liu2020prototype,ziko2020laplacian,veilleux2021realistic,TianICCV2023,HaoCVPR2023,TaoCVPR2023,TrostenCVPR2023}, to list a few. These transductive few-shot classifiers were shown to significantly outperform their inductive counterparts, with benchmarks indicating up to a 10$\%$ increase in classification accuracy \cite{boudiaf2020information}. In fact, this is in line with well-established theoretical facts in the classical literature on transductive learning \cite{vapnik1999overview,joachims1999transductive}, which points to transductive prediction as a way to alleviate the scarcity of labeled data. Importantly, and beyond theoretical justification, the transductive setting is highly relevant in a breadth of practical computer vision scenarios, in which test data may come in mini-batches. This is the case, for instance, of online video streams and various types of time-series imaging, of portable-device photos, or of pixel-level tasks such as segmentation.

In this study, we take a close look at the transductive zero-shot and few-shot inference problems for the popular vision-language pre-trained CLIP model. We first make the surprising 
observation that standard clustering models, in the zero-shot case, and recent transductive methods, in the few-shot setting, do not bring improvements comparable to those observed with 
vision-only models, scoring even below their inductive counterparts; see Tables \ref{table:zero-shot} and \ref{table:few-shot}. This might explain why the transductive setting, despite its 
popularity, has not been explored so far for vision-language models. Potential questions that may fill this gap are (i) How to build informative text-image features for transductive 
inference, leveraging the textual knowledge in vision-language models? and (ii) Are the statistical assumptions underlying standard clustering and transductive inference methods 
appropriate for text-image features? In light of these challenges, this paper brings the following contributions:
\begin{enumerate}
    \item We propose a methodology to compute text-vision probability feature vectors, setting the stage for transductive few-shot classification specifically tailored for CLIP.
    \item We reformulate the transductive zero-shot and few-shot classification challenge as an optimization problem on the unit simplex set by modeling the data with Dirichlet probability distributions. Crucially, the non-trivial deployment of the Dirichlet distributions brings substantial improvements in comparison to the common statistical models underlying standard clustering and transductive few-shot methods (e.g. Gaussian).    
    \item We propose a novel block Majorization-Minimization algorithm that addresses our problem efficiently and effectively, removing the need for cumbersome inner iterations in estimating the Dirichlet parameters.
    \item We report comprehensive evaluations, comparisons and ablations over 11 datasets, which point to the benefits of our mini-batch inference approach. On zero-shot ImageNet tasks with batches of 75 samples, the proposed method scores near $20\%$ higher than inductive zero-shot CLIP in classification accuracy. Additionally, we outperform state-of-the-art methods in the few-shot setting.

\end{enumerate}


\section{Related works}
\label{sec:related_works}

\begin{figure*}[h]
    \centering
    \includegraphics[scale=0.9]{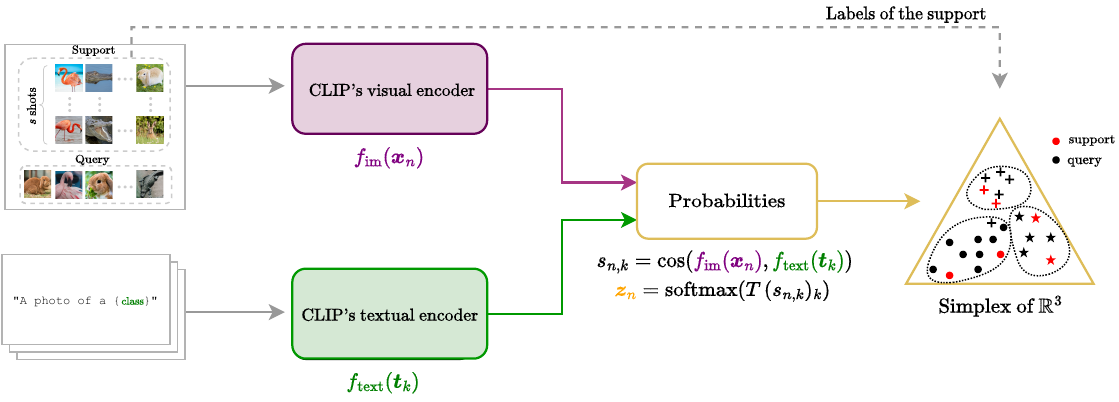}
    \caption{
    Given a transductive few-shot task, both visual and textual information are extracted from the images and class-wise prompts. The embeddings are next combined into vision-text probability vectors. Classification is carried out on the simplex set of $\R^K$ using the labels of the support set. An empty support set corresponds the the zero-shot scenario, which is akin to a clustering problem.}
    \label{fig:framework}
\end{figure*}

\subsection{Vision-language models}

Vision-Language models, like CLIP, integrate visual and textual data to improve accuracy over various vision tasks. CLIP uses a dual-encoder structure, with one deep network dedicated for image encoding and another one specilaized for text. This structure, along with proper projections at its bottleneck, yield image and text embeddings lying in the same low-dimensional vector space. Trained on a large dataset of 400 million text-image pairs, CLIP maximizes the cosine similarity between text and image embeddings using a contrastive loss. 
CLIP is pre-trained to match images with text descriptions, making it well-suited for zero-shot prediction. At inference time, to classify an image $\bs{x}$ among $K$ classes, the model predicts the class by choosing the one with the highest cosine similarity: 
\begin{equation}\label{eq:clip_zero_shot_pred}
\underset{k \in \{1, \dots, K\}}{\text{argmax}} \, \cos \left(f_{\text{im}}(\bs{x}),  f_{\text{text}}(\bs{t}_k) \right),
\end{equation}
where $f_{\text{im}}$ and $f_{\text{text}}$ are, respectively, the image and text encoders, and each $\bs{t}_k$ is based on a text prompt, typically ``a photo of a [name of class k]''.

\subsection{Few-shot classification}

\paragraph{Inductive v.s. transductive setting}
Few-shot image classification with pre-trained vision models has been the subject of extensive research recently \cite{ziko2020laplacian,chen2019closer,tian2020rethinking}. Within this area, the problem is tackled either in the \emph{transductive} or \emph{inductive} setting. The latter assumes that each instance in the testing batch is classified independently, omitting the correlations or shared information among instances~\cite{chen2019closer,wang2019simpleshot,hu2022pushing}. In contrast, transductive inference is more comprehensive, as it makes joint predictions for the entire mini-batch of query samples, leveraging their statistics and shared information. Recent research has increasingly focused on transductive few-shot learning, including, for instance, methods based on constrained clustering \cite{martin2022towards, BoudiafCVPR2023}, label propagation \cite{HaoCVPR2023, liu2018learning}, optimal transport \cite{TianICCV2023,lazarou2021iterative}, information maximization \cite{boudiaf2020information, veilleux2021realistic}, prototype rectification \cite{liu2020prototype}, among other recent approaches \cite{TaoCVPR2023,TrostenCVPR2023}. It has been consistently observed in this body of literature that the gap in accuracy between transductive and inductive methods could be considerable.

\paragraph{Few-shot CLIP} Beyond its zero-shot capabilities, the CLIP model has also been explored for few-shot image classification. In \cite{radford2021learning}, the authors evaluated linear probe, which performs a simple fine-tuning of the visual encoder's final layer using a few-shot support set (i.e., a few labeled samples in the downstream task). This approach has proven to be relatively ineffective in few-shot scenarios. Since then, a recent body of works have explored CLIP’s few-shot generalization. For instance, there is a noticeable emergence of {\em prompt learning} methods in computer vision, focusing on this specific problem \cite{zhou2022conditional,zhou2022learning,chen2022prompt}. Inspired by 
intensive recent prompt learning research in NLP \cite{ShinEMNLP2020,Hong2021NAACL}, these methods fine-tune learnable input text tokens using the few-shot support set. A different type of approaches, coined {\em adapters} \cite{gao2023clip,zhang2022tip}, fine-tune the encoded features rather than input text. For example, CLIP-Adapter \cite{gao2023clip} incorporates additional bottleneck layers to learn new features, while performing residual-style blending with the original pre-trained features. In a similar spirit, TIP-Adapter \cite{zhang2022tip} balances two prediction terms, one summarizing adaptively the information from the support set and the other preserving the textual knowledge from CLIP. All of these recent methods belong to the inductive family. To the best of our knowledge, our work is the first to explore transduction for CLIP's few-shot image classification.
\section{Proposed method }
\label{sec:method}

Throughout the paper, we define and employ specific notations to describe a single, randomly sampled few-shot task, which, during transductive prediction, is treated independently
of the other randomly sampled tasks:
\begin{itemize}
    \item $N$ is the number of images in each randomly sampled task, with $(\bs{x}_n)_{1 \leq n \leq N}$ denoting the set of images. 
    \item $K$ is the total number of distinct classes in the whole data set, among which a much smaller set of randomly sampled classes might appear in each mini-batch task, and 
    might differ from one batch to another. Hence, apart from knowing the set of $K$ classes in the whole data, as in standard inductive inference \cite{radford2021learning, zhou2022conditional, zhang2022tip}, our transductive setting do not assume any additional knowledge about the particular set of classes 
    that might appear randomly in each mini-batch.  
    \item $\support \subset \{1, \dots, N\}$ indicates the indices of samples within the support set in the few-shot setting. For all $n\in \support$, one has access to the one-hot-encoded labels $\bs{y}_n \in \{0,1\}^K$, such that for all $k \in \{1, \dots, K\}$, $y_{n,k}=1$ if $\bs{x}_n$ is an instance of class $k$, $y_{n,k}=0$ otherwise.
    \item $\query = \{1, \dots, N\} \setminus  \support$ represents the indices of samples in the query min-batch set. In our experiments, the mini-batch size $|\query|$ is set to $75$.
\end{itemize}
The goal is to predict the classes of the query samples leveraging the supervision available from the support set. Note that when the support set is empty ($\support = \varnothing$), we encounter a zero-shot scenario, which is akin to a clustering problem.

\subsection{Computing informative feature vectors}\label{sec:probability_features}

A seemingly intuitive approach to tackle the transductive challenge might be to use the visual embeddings obtained from CLIP's visual encoder as the input features for the classifier. This is analogous to CLIP's linear probe when it operates inductively.
We pinpoint two main difficulties raised by this approach:

\begin{enumerate}
    \item \textbf{Overlooking textual information}: A significant limitation of this method is that it omits the model'stextual knowledge. This is problematic as textual information is one of CLIP's most powerful features. 
    \item \textbf{Normalization dilemma}: CLIP's pre-training maximizes the scalar product between normalized textual and visual embeddings. 
Using normalized embeddings can introduce complexities in data distribution modeling, which, if misjudged, can impact the method interpretability and accuracy.
\end{enumerate}
While some works in the classification literature have explored spherical distributions like the Von Mises-Fisher \cite{hasnat2017mises,rossi2022mixture,scott2021mises} and the Fisher-Bingham \cite{hamsici2007spherical,ali2019parametric}, our approach differs to address both issues mentioned above.

Our strategy consists in defining, for every $n \in \{1, \dots, N\}$, the feature vector for the data sample $\boldsymbol{x}_n$ as CLIP's zero-shot probability. Precisely, we define
\begin{equation}\label{eq:softmax_features}
\boldsymbol{z}_n = \text{softmax}\left\{ T \, \cos\left(f_{\text{im}}(\bs{x}_n), f_{\text{text}}(\bs{t}_k)\right)_{1 \leq k \leq K} \right\} ,
\end{equation}
where $T > 0$ is a temperature parameter. Through this, both visual and textual information are incorporated into the feature vectors. Consequently, the task becomes 
a classification problem on the unit simplex of $\mathbb{R}^K$, defined as 
\begin{equation}\label{eq:def_simplex}
        \Delta_K = \left\{\boldsymbol{p}=(p_k)_{1\leq k \leq K} \in \R_{+}^K \,\Big| \, \sum_{k=1}^K p_k = 1 \right\}
\end{equation}
Observe that, for datasets with a modest number of classes, defining feature vectors according to \eqref{eq:softmax_features} also acts as a dimensionality reduction, with embedding's dimension going from $1024$ (from CLIP's ResNet50) down to $K$, the number of classes. A recap of our framework is given in Figure \ref{fig:framework}.

\subsection{Data distribution}

Given feature vectors lying within the unit simplex set of $\R^K$, we advocate modeling the data using Dirichlet distributions. The Dirichlet distribution extends the beta distribution into higher dimensions, serving as a natural choice for modeling probability vectors over the simplex. 
For each class $k$ within the set $\{1, \dots, K\}$, the data is assumed to follow a Dirichlet distribution, characterized by positive parameters $ \boldsymbol{\alpha}_k = (\alpha_{k,i})_{1 \leq i \leq K} \in (0, +\infty)^K$, which describes the distribution shape. An illustration in $\R^3$ is given is Appendix \ref{app:dirichlet_plot}. Mathematically, the density function is given by, for every $\boldsymbol{z}=(z_i)_{1\leq i \leq K} \in \mathbb{R}^K$,
\begin{equation}
\label{eq:dirichlet_distribution}
\mathrm{p} \left( \boldsymbol{z} ~|~ \boldsymbol{\alpha}_k \right ) = \frac{1}{\mathcal{B}(\boldsymbol{\alpha}_k)} \prod_{i=1}^K z_{i}^{\alpha_{k, i} -1} \, \mathbbm{1}_{\boldsymbol{z} \in \Delta_K},
\end{equation}
where normalization factor $\mathcal{B}(\boldsymbol{\alpha}_k)$ is expressed as
\begin{equation}
   \mathcal{B}(\boldsymbol{\alpha}_k) = \frac{\prod_{i=1}^K \Gamma(\alpha_{k,i}) }{\Gamma\left(\sum_{i=1}^K \alpha_{k,i}\right)}, 
\end{equation}
and $\Gamma$ denoting the Gamma function.

\subsection{Simplex-based classification criterion}
The proposed method simultaneously determines: (\textit{i}) the soft assignment vectors $\bs{u} = (\bs{u}_n)_{1 \leq n \leq N}$ within the simplex $(\Delta_K)^N$, where the $k$-th component $u_{n, k}$ of vector $\bs{u}_n$  specifies the probability for the $n$-th sample belonging to class $k$; 
(\textit{ii}) the Dirichlet distribution parameters $\bs{\alpha} =(\bs{\alpha}_k)_{1 \leq k \leq K}$ where each $\bs{\alpha}_k$ is a $K$-dimensional vector with nonnegative components. 
We achieve this through the following maximum-likelihood estimation
\begin{alignat}{2}\label{eq:regularized_problem}
    &\underset{\bs{u}, \bs{\alpha}}{\mathrm{minimize}}\,   &&- \mathcal{L}(\bs{u}, \bs{\alpha}) + \Phi(\bs{u})+ \lambda  \Psi(\bs{u}),\\
    &\text{subject to} && \, \bs{u}_n \in \Delta_K \quad \forall n \in \query, \nonumber\\
    & && \, u_{n,k} = y_{n, k} \quad \forall n \in \support, \, \forall k \in \{1, \dots, K\}. \nonumber
\end{alignat}
In \eqref{eq:regularized_problem}, $\mathcal{L}$ is the log-likelihood model fitting objective for clustering:
\begin{equation}\label{eq:likelihood_dirichlet}
    \mathcal{L}(\bs{u}, \bs{\alpha})= \sum_{n = 1}^N \sum_{k=1}^K u_{n,k} \ln( \p \left( \boldsymbol{z}_{n} ~|~ {\boldsymbol{\alpha}_k} \right )),
\end{equation}
where the density functions are defined by the Dirichlet models in \eqref{eq:dirichlet_distribution}. When the support set is not empty, this term also includes the supervision derived from the labeled instances.
Term $\Phi$ acts as a barrier imposing the nonnegativity constraints on assignment variables, as in the soft $K$-means objective \cite[p.289]{mackay2003information}, and is defined as
\begin{equation}\label{eq:regularization_barrier}
  \Phi(\bs{u})= \sum_{n=1}^N \sum_{k=1}^K u_{n,k}\ln u_{n,k}.
\end{equation}
Finally, the penalty function $\Psi$, weighted by parameter $\lambda \in [0, +\infty)$, evaluates a partition complexity \cite{martin2022towards,Boykov-ICCV-05}, linked to the Minimum Description Length (MDL) 
concept in information theory:
\begin{equation}\label{eq:regularization_mdl}
  \Psi(\bs{u})=  - \sum_{k=1}^K \pi_k \ln \pi_k,
\end{equation}
where $\pi_k= \frac{1}{|\query|} \sum_{n\in \query} u_{n, k}$ is the proportion of query samples within class $k$. This MDL term penalizes the number of non-empty clusters,
encouraging low-complexity partitions, i.e., with lower numbers of clusters.

\section{Proposed algorithm}\label{sec:algorithm}

To tackle the minimization problem \eqref{eq:regularized_problem}, our algorithm alternates minimization steps on the assignment variables and the Dirichlet parameters, producing sequences $(\bs{u}^{(\ell)})_{\ell \in \N}$ and $(\bs{\alpha}^{(\ell)})_{\ell \in \N}$ making the objective function decrease. 

\subsection{Minimization step w.r.t Dirichlet parameter}\label{sec:update_alpha}

Suppose $\bs{u} \in \Delta_K^N$ is fixed. The estimation step with respect to the Dirichlet parameters consists in maximizing the log-likelihood in \eqref{eq:likelihood_dirichlet}. Given the separability of the cost with respect to the variables $(\sbmm{\alpha}_k)_{1 \leq k \leq K}$, we can, without loss of generality, consider the minimization of the function $F_k$ defined as $(\forall \sbmm{\alpha}_k\in (0, +\infty)^K) $
\begin{multline}\label{eq:def_negative_log_likelihood_k}
    F_k(\sbmm{\alpha}_k) = \sum_{n = 1}^N u_{n,k} \Bigg(\sum_{i=1}^K -(\alpha_{k,i} -1) \ln z_{n,i} \\ + \sum_{i=1}^K \ln \Gamma(\alpha_{k,i}) - \ln \Gamma\left(\sum_{i=1}^K \alpha_{k,i}\right) \Bigg).
\end{multline}

The minimization of the Dirichlet negative log-likelihood \eqref{eq:def_negative_log_likelihood_k} has already been explored in the literature \cite{narayanan1991algorithm,huang2005maximum,minka2000estimating}. The main strategy consists in resorting to a Majorization-Minimization (MM) algorithm. Specifically, a generic MM procedure corresponds to finding a minimizer $\bs{\alpha}_k^*$ of $F_k$ by iteratively producing a sequence $(\bs{\alpha}_k^{(m)})_{m \in \N}$ such that, for every $m \in \N$,
\begin{equation}\label{eq:MM_scheme}
    \bs{\alpha}_k^{(m+1)}  = \underset{\bs{\alpha}_k \in (0, +\infty)^K}{\text{argmin}}\; q(\bs{\alpha}_k; \bs{\alpha}_k^{(m)}),
\end{equation}
where for all $\bs{\beta}_k \in (0, +\infty)^K$, $q(\,\cdot\,; \bs{\beta}_k)$ is a so-called \emph{tangent majorant} of $F_k$, satisfying 
\begin{equation}
    \syst{F(\bs{\alpha}_k)  \leq q(\bs{\alpha}_k; \bs{\beta}_k),  \\
    F(\bs{\beta}_k) =q(\bs{\beta}_k; \bs{\beta}_k).}
\end{equation}
The efficiency of the procedure \eqref{eq:MM_scheme} is highly dependent on the choice of the majorant. In \cite{minka2000estimating}, the author proposed a majorant function of \eqref{eq:def_negative_log_likelihood_k} which consists in linearizing the concave term $\bs{\alpha}_k \mapsto -\ln \Gamma \left(\sum_{i=1}^K \alpha_{k,i} \right)$ at $\bs{\beta}_k$. The resulting MM algorithm was used for simplex clustering in \cite{blei2003latent,pal2022clustering}. 
However, minimizing this majorant requires inverting the digamma function (i.e., the derivative of the log-Gamma function) with a Newton method, which can jeopardize the numerical convergence and slow down the overall algorithm. 

In the following lemma, we introduce a novel tight majorant of $F_k$ which yields closed-form updates, therefore avoiding sub-iterations within the MM algorithm. 

\begin{lemma}[Majorant of the negative log-likelihood]\label{lemma:majorant_F}
Let $\varphi = \ln \Gamma( \cdot +1)$. 
Then, for any $\bs{\beta}_k = (\beta_{k, i})_{1\leq i \leq K} \in (0, +\infty)^K$, the function $q(\,\cdot\,; \bs{\beta}_k)$ defined as, for every $ \bs{\alpha}_k\in  (0, +\infty)^K)$,
\begin{align}\label{eq:majorant_F}
&q(\bs{\alpha}_k; \bs{\beta}_k)\nonumber\\
&= \sum_{n=1}^N u_{n,k}\Bigg[ \sum_{i=1}^K \Big(-(\alpha_{k, i} -1) \ln z_{n,i}
-\ln \alpha_{k, i}
\nonumber\\
&+ \varphi(\beta_{k,i})
+\varphi'(\beta_{k,i}) (\alpha_{k,i}-
\beta_{k,i})
+\frac{c(\beta_{k,i})}{2}
(\alpha_{k,i}-
\beta_{k,i})^2\Big)
\nonumber\\ 
&
-\ln \Gamma\Big(\sum_{i=1}^K \beta_{k,i}\Big)
-  \left( \sum_{i=1}^K (\alpha_{k,i}- \beta_{k,i})\right)(\ln \Gamma)'\Big(\sum_{i=1}^K \beta_{k,i}\Big)\Bigg]
\end{align}
is a tangent majorant of $F_k$ at $\bs{\beta}_k$, where the function $c$ is defined by
\begin{equation}\label{e:defcourb}
        c\colon t  \mapsto \begin{cases}
        \varphi''(0) & \mbox{if $t = 0$,}\\
        \displaystyle 2\frac{\varphi(0)-\varphi(t)+\varphi'(t)t}{t^2} & \mbox{otherwise.}
        \end{cases}
    \end{equation}
\end{lemma}
A proof of Lemma \ref{lemma:majorant_F} is provided in Appendix \ref{app:majorant_F}. At each iteration of our MM procedure, the minimizer of the majorizing function \eqref{eq:majorant_F} is the positive root of a quadratic polynomial equation, resulting in Algorithm \ref{algo:MM_update_alpha}. In Appendix \ref{app:majorant_F}, we show that this majorant speeds up the MM scheme over Minka's \cite{minka2000estimating}.



 \begin{center}
\RestyleAlgo{ruled}
	\begin{algorithm*} 
Initialize $\boldsymbol{\alpha}_k^{(0)} = \bs{\alpha}_k$.\\
\For{$m = 0, 1, \ldots,$}
{\For{$i \in \{1, \dots, K\}$}{
$\displaystyle
b_{k,i} = 
\varphi'(\alpha_{k, i}^{(m)}) - (\ln\Gamma)'\left(\sum_{j=1}^K \alpha_{k, j}^{(m)} \right)
-c(\alpha_{k, i}^{(m)}) \alpha_{k, i}^{(m)}
-\left(\sum_{n=1}^N u_{n, k}\right)^{-1} \sum_{n=1}^N u_{n, k} \ln (z_{n,i}).$\\
$\displaystyle
\alpha_{k, i}^{(m+1)}
= \left(-b_{k,i}+
\sqrt{b_{k,i}^2
+ 4 c(\alpha_{k,i}^{(m)})} \right)\Big/
2c(\alpha_{k,i}^{(m)}).$
}}
\caption{MM-quadratic($\bs{u}_{\cdot, k}$, $\bs{\alpha}_k$) \label{algo:MM_update_alpha}}
\end{algorithm*}
\end{center}

\subsection{Minimization step w.r.t assignment variable}\label{sec:update_u}
Let the iteration number $\ell \in \N$ and $\bs{\alpha} = (\bs{\alpha}_k)_{1 \leq k \leq K} \in ((0, +\infty)^K)^K$ be fixed.
Because of the partition complexity term in \eqref{eq:regularization_mdl}, the direct minimization of the partial function with respect to $\bs{u}_n$, for every $n \in \query $, is not closed form. Since the partition complexity penalty is concave, we propose to replace it by its linear upper-bound, leading us to minimize
\begin{multline}\label{eq:cost_function_wrt_u}
  \bs{u}_n \mapsto -\sum_{k=1}^K u_{n,k} \ln( \p \left( \boldsymbol{z}_{n} ~|~ {\boldsymbol{\alpha}_k} \right )) +\sum_{k=1}^K u_{n,k}\ln u_{n,k}\\ -\frac{\lambda}{|\query|}(\ln(\bs{\pi}^{(\ell+1)})+1)^\top( \bs{u}_n  - \bs{u}_n^{(\ell)}) ,
\end{multline}
under the simplex and supervision constraints. In \eqref{eq:cost_function_wrt_u}, $\bs{\pi}^{(\ell+1)} = (\pi_k^{(\ell+1)})_k$ 
is the vector whose $k$-th component is  
$\pi_k^{(\ell+1)}= \frac{1}{|\query|} \sum_{n\in \query} u_{n, k}^{(\ell)}$, 
and the log function operates componentwise.

Solving this minimization problem yields the updates, for every $n \in \query$,
\begin{equation*}
   \bs{u}_n^{(\ell +1)} =  \text{\normalfont softmax}\left(  \left(\ln \p \left( \boldsymbol{z}_{n} ~|~ {\boldsymbol{\alpha}_k}  \right)+ \frac{\lambda}{|\query|} \ln(\pi_{k}^{(\ell+1)}) \right)_{k}  \right).
\end{equation*}
Details for deriving this expression are given in Appendix~\ref{app:details_step_u}.

\subsection{Global algorithm and class-assignment}\label{sec:class_cluster_matching}
Finally, given the estimation steps on the assignment variables $\bs{u}=(\bs{u}_n)_{n \in \query}$ and on the Dirichlet parameters $(\bs{\alpha}_k)_{1 \leq k \leq K}$ derived respectively in Sections \ref{sec:update_u} and \ref{sec:update_alpha}, our complete procedure to tackle the minimization problem in \eqref{eq:regularized_problem} is detailed in Algorithm \ref{algo:em_dirichlet}. We name it EM-Dirichlet as it shares close links to the EM algorithm, as it will be established in Proposition \ref{prop:links_EM} in Section \ref{sec:links_clustering}.

In the zero-shot scenario, the tasks at hand can be seen as a form of simplex clustering. There exists a straightforward method to map each cluster to a corresponding class label in an injective manner. Let $(\mathcal{C}_k)_{k\in \mathcal{K}}$ denote the set of non-empty clusters found with a clustering method, for instance ours, with $\mathcal{K}$ a subset of $k \in \{1, \dots, K\}$ and for all $k \in \mathcal{K}$, $\mathcal{C}_k$ a subset of $\query$. We proceed in the following way:

\begin{enumerate}
    \item For each $k\in \mathcal{K}$, calculate the mean of cluster $k$, $\boldsymbol{m}_k = (m_{k,\ell})_{1 \leq \ell \leq K} \in \Delta_K$, as $
        \boldsymbol{m}_k = \frac{1}{|\mathcal{C}_k|} \sum_{n \in \mathcal{C}_k} \boldsymbol{z}_n.$
    The element $m_{k,\ell}$ is interpreted as the probability that cluster $k$ is associated with class $\ell$. While it may seem intuitive to assign cluster $k$ the class $\ell$ for which $m_{k,\ell}$ is maximal, this could lead to multiple clusters being assigned to the same class, which we wish to avoid.

    \item Resolve the class-to-cluster assignments through a bipartite graph matching that maximizes $
    \sum_{k\in \mathcal{K}} \sum_{\ell=1}^K a_{k, \ell} m_{k, \ell}$
    over all possible assignment matrices $\boldsymbol{A}=(a_{k, \ell})\in \{0,1\}^{|\mathcal{K}| \times K}$ that satisfy $\boldsymbol{A}^\mathsf{T} \mathbf{1}_{|\mathcal{K}|}= \mathbf{1}_K.$
    This class assignment integer linear programming problem can be solved with algorithms such as \cite{crouse2016implementing}.
    An illustration of this process can be found in Appendix \ref{app:class_assignment}.
\end{enumerate}

 \begin{center}
\RestyleAlgo{ruled}
	\begin{algorithm*} 
Initialize $\boldsymbol{u}^{(0)}$ as CLIP's probabilities and for all $k\in \{1, \dots, N\}$, $\bs{\alpha}_k^{(0)}=\mathbf{1}_{K}$.\\
\For{$\ell = 0, 1, \ldots,$}
{{\color{blue}\tcp{Update Dirichlet parameter for each class}}
$\displaystyle\bs{\alpha}_k^{(\ell +1)} = \text{MM-quadratic}(\bs{u}_{\cdot, k}^{(\ell)}, \bs{\alpha}_k^{(\ell)})$,  \quad $\forall k \in\{1, \dots, K\},$\\
{\color{blue}\tcp{Update class proportions}}
$\displaystyle\pi_{k}^{(\ell+1)} = \frac{1}{|\query|} \sum_{n\in \query} u_{n, k}^{(\ell)}$,  \quad $\forall k \in\{1, \dots, K\},$\\
{\color{blue}\tcp{Update assignment variable for all query samples}}
$\displaystyle\bs{u}_n^{(\ell +1)} =  \text{\normalfont softmax}\left(  \left(\ln \p \left( \boldsymbol{z}_{n} ~|~ {\boldsymbol{\alpha}_k^{(\ell+1)}}  \right)+ \frac{\lambda}{|\query|} \ln(\pi_{k}^{(\ell+1)}) \right)_{ k}  \right) $, \quad $\forall n \in \query.$ 
}
\caption{EM-Dirichlet \label{algo:em_dirichlet}}
\end{algorithm*}
\end{center}

\section{Links with other clustering and transductive few-shot objectives}\label{sec:links_clustering}

The general log-likelihood model fitting objective in \eqref{eq:likelihood_dirichlet}, also referred to as probabilistic $K$-means \cite{Kearns-UAI-97,Boykov-ICCV-05}, is well-established in the clustering literature. Indeed, it is a generalization of the ubiquitous $K$-means, which corresponds to the particular choice of the Gaussian distribution for the densities in \eqref{eq:likelihood_dirichlet}, with covariance matrices fixed to the identity matrix. This general objective has a strong, inherent bias towards $K$-balanced partitions, a theoretically well-established fact in the clustering literature \cite{Kearns-UAI-97,Boykov-ICCV-05}. To mitigate this bias and address realistic, potentially imbalanced few-shot query sets, the recent transductive few-shot method in \cite{martin2022towards} coupled the MDL term in (\ref{eq:regularization_mdl}) with the standard $K$-means objective. This corresponds to the general data-fitting function we tackle in (\ref{eq:likelihood_dirichlet}), but with the likelihood densities assumed to be Gaussian. As we will see in our experiments (Table \ref{table:few-shot}), the non-trivial deployment of the Dirichlet model is crucial, outperforming significantly~\cite{martin2022towards} in CLIP's few-shot setting. Furthermore, we show in the following an interesting result, which connects the general unbiased clustering problem we propose in (\ref{eq:regularized_problem}), to the well-known Expectation-Maximization (EM) algorithm for mixture models ~\cite[p.438]{bishop06PRML}. Indeed, optimizing the objective in (\ref{eq:regularized_problem}) could be viewed as a generalization of EM, enabling to control the class-balance parameter $\lambda$. 
\begin{proposition}\label{prop:links_EM}
Consider the unsupervised classification problem, i.e. $\support = \varnothing$.
    Suppose the value of $\lambda$ in \eqref{eq:regularized_problem} is set to the size of the query set, i.e., $\lambda = |\query|$. Then Algorithm \ref{algo:em_dirichlet} is equivalent to the EM algorithm when applied to a generic mixture model
 \begin{equation} 
\label{eq:dirichlet_mixture}
\mathrm{p} \left( \boldsymbol{z}_{n} ~|~ \boldsymbol{\pi}, \boldsymbol{\alpha} \right ) = \sum_{k=1}^K {\pi_k} \mathrm{p} \left( \boldsymbol{z}_{n} ~|~ {\boldsymbol{\alpha}_k} \right ),
\end{equation}
where $\boldsymbol{\pi} = (\pi_k)_{1 \leq k \leq K} \in \Delta_K$ are the mixture coefficients.
\end{proposition}
The proof of Proposition \ref{prop:links_EM} is given in Appendix \ref{app:links_EM}.
\section{Experiments}
\label{sec:exp}

We evaluated our method on 11 publicly accessible image classification datasets which were also utilized in CLIP \cite{radford2021learning}: ImageNet \cite{russakovsky2015imagenet}, Caltech101 \cite{fei2006one}, OxfordPets \cite{parkhi12cats}, StanfordCars \cite{krause20133object}, Flowers102 \cite{nilsback2008automated}, Food101 \cite{bossard2014food}, FGVCAircraft \cite{maji2013fine}, SUN397 \cite{xiao2010sun}, DTD \cite{cimpoi14describing}, EuroSAT \cite{helber2019eurosat} and UCF101 \cite{soomro2012ucf101}. 
To ensure reproducibility, we adhere to the dataset splits provided by CoOp \cite{zhou2022conditional} and use the prompts employed in TIP-Adapter \cite{zhang2022tip}.  All experiments are conducted using CLIP's pre-trained ResNet50 visual encoder. The temperature in the probabilities \eqref{eq:softmax_features} is fixed to $T=30$.

\subsection{Zero-shot}\label{sec:zero_shot_generation}

\paragraph{Tasks generation}
For generating query sets in our transductive zero-shot setting, we employ a practical approach that maintains manageable batch sizes. At each new task (mini-batch), we randomly select the classes that will be represented in the query set, with the actual number of distinct classes ranging from 3 to 10, also selected at random. It is important to note that the set of classes occurring in each batch remain undisclosed, and vary randomly from one batch to another, ensuring that the clustering task is still performed over all $K$ potential classes present in the whole dataset. Subsequently, we randomly select $|\query|=75$ images in to the chosen classes to constitute the query set. During transductive inference, the query set of each task is treated independently of the other randomly sampled tasks.

\paragraph{Comparative methods}
We conduct a comparative evaluation of our clustering methodology, EM-Dirichlet, and its variant utilizing hard assignments, denoted as Hard EM-Dirichlet, against a range of clustering objective functions and algorithms: Hard and soft $K$-means~\cite[p.286]{mackay2003information}, EM for Gaussian mixtures with identity covariance (EM-Gaussian (cov. Id)) and with diagonal covariance (EM-Gaussian (cov. diag))~\cite[p.438]{bishop06PRML}, and Hard KL $K$-Means \cite{cao2013sail}. Furthermore, our comparison provides a full ablation study of the terms in general objective function (\ref{eq:regularized_problem}):
\begin{enumerate}
    \item The log-likelihood model fitting term
\eqref{eq:likelihood_dirichlet}, which varies across Gaussian (employed in Hard $K$-means, Soft $K$-means, EM-Gaussian), 
    and Dirichlet (in our method).
    \item The entropic barrier (\ref{eq:regularization_barrier}) featured in both Soft $K$-means and the EM-based approaches.
    \item The MDL partition-complexity term (\ref{eq:regularization_mdl}), incorporated exclusively in the EM methods.
\end{enumerate}
Initialization is uniform across different clustering techniques, utilizing CLIP's predictions from Equation~\eqref{eq:softmax_features}. In all EM-based methods, the regularization parameter $\lambda$ is set according to $\lambda = \frac{5}{K}|\query|$, to maintain consistency across comparisons.

\paragraph{Results}
We assess the clustering methods on zero-shot tasks, using both the visual embeddings and the combined text-vision feature vectors. We also include the zero-shot classification results from CLIP. In Table \ref{table:zero-shot}, we report average accuracy over 1,000 tasks using the graph cluster-to-classes assignment described in Section \ref{sec:class_cluster_matching}. Table \ref{table:zero-shot} conveys several crucial messages:
\begin{itemize}
\item Clustering visual embeddings alone does not suffice to surpass inductive CLIP's zero-shot performance. Incorporating textual information via probability features enhances the performance, even for methods initially designed for Gaussian distributions.
\item Gaussian-based data-fitting approaches are sub-optimal for simplex clustering. Replacing the Gaussian metric with a Kullback-Leibler divergence
is beneficial. Employing a Dirichlet data-fitting term within the EM framework significantly improves the results compared to EM-Gaussian methods, highlighting the necessity of accurate data distribution modeling.
\item Introducing the partition complexity term (in the EM methods), which discourages overly balanced predictions, proves advantageous for the performance.
\item Using an adapted transductive model like Hard EM-Dirichlet, accuracy improves considerably, showing a $9\%$ rise across 11 datasets, and nearly $20\%$ on ImageNet.
\end{itemize}

In Appendix \ref{app:variation_query_size}, we show that zero-shot performance improves with larger query set sizes, indicating enhanced transduction efficiency with increasing mini-batch size.

\begin{table*}
  \centering
    \begin{tabular}{ll|ccccccccccc|c}
                       &  & \rotatebox[origin=l]{90}{Food101} & \rotatebox[origin=l]{90}{EuroSAT} &  \rotatebox[origin=l]{90}{DTD} &\rotatebox[origin=l]{90}{OxfordPets}  & \rotatebox[origin=l]{90}{Flowers102}  & \rotatebox[origin=l]{90}{Caltech101}  & \rotatebox[origin=l]{90}{UCF101}  & \rotatebox[origin=l]{90}{FGVC Aircraft} &\rotatebox[origin=l]{90}{Stanford Cars} &\rotatebox[origin=l]{90}{SUN397} &\rotatebox[origin=l]{90}{ImageNet} &\rotatebox[origin=l]{90}{\textbf{Average}} \\ \hline
    &     Zero-shot CLIP & 77.1 & 36.5 & 42.9 & 85.1 & 66.1 & 84.4 & 61.7 & 17.1 & 55.8 & 58.6 & 58.3 & 58.5\\ \hline
    \multirow{4}{*}{\rotatebox[origin=c]{90}{\textcolor{blue}{Vis. embs.}}} & {Hard K-means}    & 52.2 & 37.9 & 40.0 & 54.5 & 44.8 & 62.9 & 49.2 & 14.3 & 22.4 & 39.6 & 29.3 & 40.6\\
    & {Soft K-means}                                                                 & 17.6 & 29.9 & 19.1 & 40.4 & 36.1 & 21.3 & 13.3 & 10.6 & 11.1 & 9.1 & 10.1 & 19.8\\
    & {EM-Gaussian ($\mathrm{Id}$ cov.)}                                                   & 14.0 & 14.5 & 9.4 & 6.9 & 5.3 & 30.3 & 7.4 & 1.9 & 2.5 & 5.3 & 3.9 & 8.3\\  
    & {EM-Gaussian (diag cov.)}                                                   & 51.4 & \textbf{40.6} & 37.5 & 59.2 & 45.6 & 61.8 & 47.2 & 13.6 & 24.3 & 35.1 & 28.3 & 40.4\\ \hline 

    \multirow{6}{*}{\rotatebox[origin=l]{90}{ \textcolor{blue}{Probabilities}}} & {Hard K-means}       & 49.5 & 35.2 & 38.7 & 62.4 & 44.5 & 52.2 & 46.6 & 14.5 & 29.6 & 41.4 & 31.0 & 40.5\\
    & {Soft K-means}                                                                 & 41.8 & 21.5 & 18.3 & 56.6 & 34.3 & 50.5 & 30.2 & 7.2 & 34.8 & 18.8 & 19.1 & 30.3\\
     & {EM-Gaussian ($\mathrm{Id}$ cov.)}                                                                & 21.4 & 14.5 & 16.5 & 21.1 & 23.1 & 33.6 & 19.3 & 6.8 & 18.5 & 18.7 & 19.1 & 19.3\\ 
     & {EM-Gaussian (diag cov.)}                                                   & 63.3 & 33.1 & 38.7 & 71.1 & 51.1 & 66.6 & 56.0 & 16.5 & 46.9 & 54.8 & 48.5 & 49.7 \\  
    & {Hard KL K-means}                                                                   & 72.2 & 34.9 & 40.8 & 73.0 & 61.1 & 72.0 & 60.6 & 17.7 & 56.2 & 61.8 & 61.0 & 55.6\\
        \rowcolor{lightsalmon!20}  & EM-Dirichlet                                & 88.2 & 33.0 & 47.7 & 87.3 & 71.5 & 88.4 & 69.0 & 19.2 & 65.5 & 77.3 & 76.9 & 65.8\\ 
       \rowcolor{lightsalmon!40}        &    Hard EM-Dirichlet                   & \textbf{90.2} & 36.1 & \textbf{49.3} & \textbf{90.9} & \textbf{73.1} & \textbf{89.7} & \textbf{70.3} & \textbf{20.4} & \textbf{67.7} & \textbf{78.5} & \textbf{77.6} & \textbf{67.6}\\ 
    \end{tabular}
    \caption{Average accuracy of clustering methods over 1,000 zero-shot classification tasks. Inference is performed both on the visual embeddings and on the text-vision probability features.}
    \label{table:zero-shot}
\end{table*}

\subsection{Few-shot}

\paragraph{Task generation}
We follow the realistic transductive few-shot evaluation protocol proposed recently in \cite{martin2022towards}. Specifically, the query sets are constructed with a fixed number of effective classes 
$k_{\text{eff}} = 5$, from which $|\query|$ samples are randomly selected. This approach aligns with established few-shot protocols in the literature \cite{snell2017prototypical, liu2018learning, veilleux2021realistic}. These classes remain undisclosed during inference, ensuring the task is a $K$-way classification. The support set is created by uniformly selecting $s$ images from each of the $K$ classes. The ensuing results are derived performing few-shot tasks with $1$, $2$, $4$, $8$, and $16$ shots. During inference on the test set, the size of the query set is set to $|\query|=75$, while for validation, the size is reduced to$|\query|=35$ due to data limitations.

\paragraph{Hyper-parameters}
Parameter $\lambda$ in EM-Dirichlet is set to the fixed value $\lambda = \frac{k_\text{eff}}{K}|\query|$. Methods with tunable hyper-parameters are fine-tuned using the validation split provided with each dataset. In line with \cite{hu2022pushing}, validation is performed on five $s$-shot tasks across all datasets and for every shot number. These tasks, crafted as previously detailed, use support and query instances drawn from the validation set. The hyper-parameters are then optimized through a grid search to maximize accuracy on the validation set. 

\begin{table*}[htb]
  \centering
   \resizebox{1.0\hsize}{!}{
    \begin{tabular}{ll|ccccccccccc|c|l}
                       & & \rotatebox[origin=l]{90}{Food101} & \rotatebox[origin=l]{90}{EuroSAT} &  \rotatebox[origin=l]{90}{DTD} &\rotatebox[origin=l]{90}{OxfordPets}  & \rotatebox[origin=l]{90}{Flowers102}  & \rotatebox[origin=l]{90}{Caltech101}  & \rotatebox[origin=l]{90}{UCF101}  & \rotatebox[origin=l]{90}{FGVC Aircraft} &\rotatebox[origin=l]{90}{Stanford Cars} &\rotatebox[origin=l]{90}{SUN397} &\rotatebox[origin=l]{90}{ImageNet} &\rotatebox[origin=l]{90}{\textbf{Average}} &\quad \rotatebox[origin=l]{90}{\textbf{Time} ($s$)}\\ \hline
        \multirow{2}{*}{\rotatebox[origin=c]{90}{\textcolor{blue}{Ind.}}} 
        & Tip-Adapter \cite{zhang2022tip}     & 76.7 & \textbf{72.5} & 54.7 & 86.4 & 83.2 & 88.8 & 72.1 & 23.7 & 63.9 & 66.7 & 62.7 & 68.3 & 6.76$\times$10$^{0}$\\
        & CoOp \cite{zhou2022learning}     & 76.3 & 63.2 & 52.2 & 86.2 & 81.0 & 87.7 & 67.0 & 22.2 & 61.3 & 63.4 & 59.9 & 65.5 & 3.35$\times$10$^{3}$\\ \hline
        \multirow{5}{*}{\rotatebox[origin=c]{90}{\textcolor{blue}{Trans.}}} 
        & BDSCPN \cite{liu2020prototype}     & 74.7 & 46.1 & 45.2 & 81.3 & 74.2 & 82.0 & 59.0 & 18.0 & 48.1 & 54.5 & 49.2 & 57.5 & 4.49$\times$10$^{-1}$ \\
        & Laplacian Shot \cite{ziko2020laplacian}     & 76.6 & 53.0 & 52.6 & 88.4 & 85.5 & 86.8 & 67.0 & 22.2 & 60.4 & 63.8 & 56.3 & 64.8 & 2.10$\times$10$^{-1}$\\
        & $\alpha$-TIM \cite{veilleux2021realistic}     & 66.1 & 46.1 & 45.3 & 87.1 & 79.1 & 83.3 & 59.4 & 20.4 & 53.4 & 53.4 & 42.7 & 57.8 & 1.65$\times$10$^{0}$\\
        & PADDLE \cite{martin2022towards}              & 71.8 & 45.9 & 50.0 & 84.7 & 82.3 & 81.9 & 63.7 & 21.3 & 56.1 & 60.6 & 52.1 & 60.9 & 4.04$\times$10$^{-1}$\\
        \rowcolor{lightsalmon!20} & EM-Dirichlet       & \textbf{88.7} & 50.8 & \textbf{62.6} & \textbf{92.5} & \textbf{91.3} & \textbf{90.1} & \textbf{76.1} & \textbf{24.9} & \textbf{73.5} & \textbf{80.9} & \textbf{78.4} & \textbf{73.6} & 1.04$\times$10$^{0}$\\
        \rowcolor{lightsalmon!40} &Hard EM-Dirichlet  & 87.9 & 50.8 & 60.5 & 91.7 & 90.5 & 89.8 & 75.3 & 24.2 & 72.6 & 80.2 & 78.3 & 72.9 & 6.97$\times$10$^{-1}$\\
    \end{tabular}}
    \caption{Evaluation of our approach against two benchmarks – 1) inductive methods specifically designed for few-shot classification using CLIP, and 2) transductive few-shot methods applied to probability feature vector classification. The analysis encompasses 1,000 distinct 4 shots tasks. We also report average execution time for a single task, computed over 1,000 tasks, on the ImageNet dataset.}
    \label{table:few-shot}
\end{table*}

\paragraph{Results}
We evaluate the accuracy of our proposed transductive methods, EM-Dirichlet and Hard EM-Dirichlet, against several recent transductive few-shot methods, including BD-CSPN \cite{liu2020prototype}, Laplacian Shot \cite{ziko2020laplacian}, $\alpha$-TIM \cite{veilleux2021realistic}, and PADDLE \cite{martin2022towards}. Additionally, we benchmark against two inductive few-shot methods designed for CLIP: TIP-Adapter \cite{zhang2022tip} and CoOp \cite{zhou2022learning}. The results, averaged across $1,000$ tasks with $4$ shots, are presented in Table \ref{table:few-shot} and for the other number of shots in Appendix \ref{app:additional_results_few_shot}.

Our method surpasses competing approaches on the majority of datasets, with a more pronounced advantage observed on challenging datasets that have a large number of classes, such as SUN397 and ImageNet. The accuracy gap between our method and the inductive ones shows the benefits of transductive inference. On the other hand, the inferior perfomance of other transductive methods can be attributed to their lack of adaptability to simplex classification.
 
Interestingly, our results indicate that on some datasets such as Food101, our method perform better in the zero-shot than in the few-shot setting. This is consistent with Radford et al. \cite{radford2021learning}, suggesting that few labeled examples can negatively impact classification, possibly due to outliers or ambiguous examples in the support set.
 
Lastly, we observe that inductive methods outperform ours on the EuroSAT dataset. This might be due to the inclusion of text information in the vision-text features. While typically advantageous, it is possible that the text information introduces a confounding effect specific to this dataset.


\section{Conclusion}
In conclusion, our study expands transductive inference to vision-language models like CLIP, previously unexplored in this domain. We demonstrate that the transductive methodology can boost image classification accuracy, including in zero-shot scenarios. Future work could apply our transductive CLIP approach to other tasks like segmentation and out-of-distribution detection.

\newpage
{
    \small
    \bibliographystyle{ieeenat_fullname}
    \bibliography{main}
}

\clearpage
\maketitlesupplementary
\setcounter{subsection}{0}
\renewcommand\thesubsection{\Alph{subsection}}

\subsection{Illustration of a Dirichlet distribution}\label{app:dirichlet_plot}

\begin{figure}[h]
    \centering
    \includegraphics[width=90pt, height=80pt]{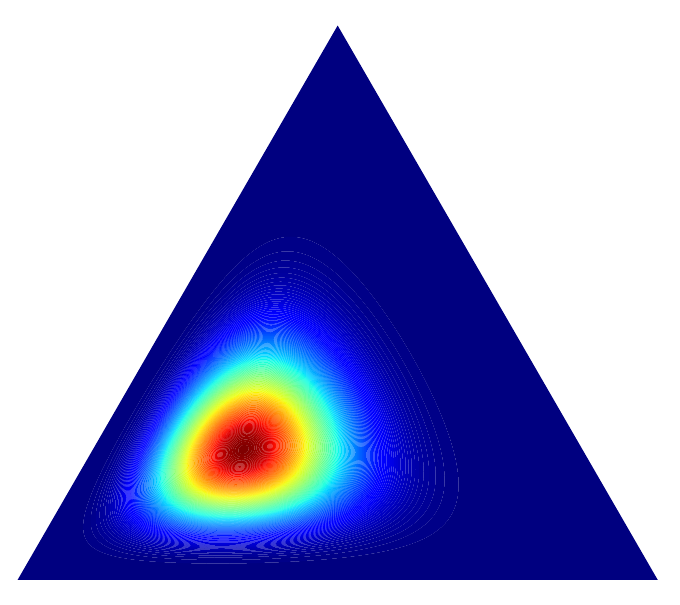}
    \includegraphics[width=120pt, height=80pt]{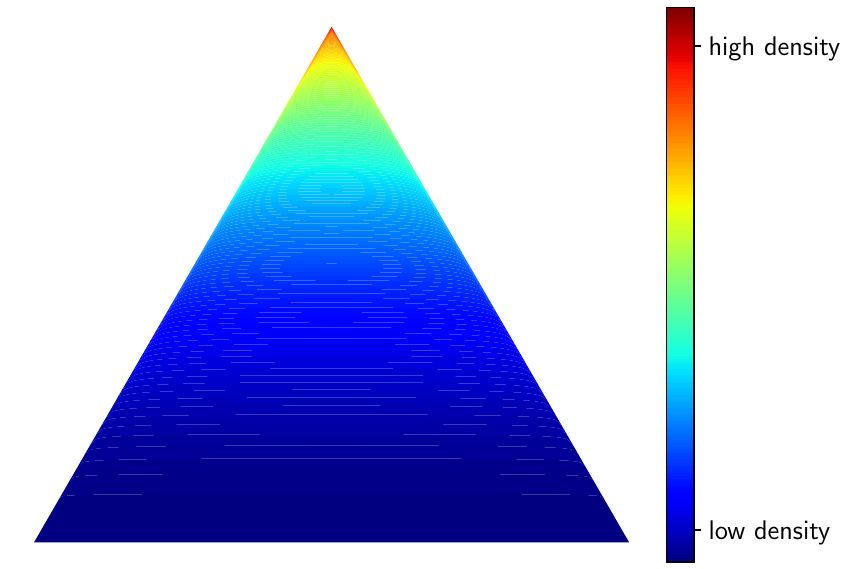}
    \caption{Examples of Dirichlet distributions on the simplex of $\R^3$, for $\bs{\alpha}=(10, 5.0, 5.0)$ (left) and $\bs{\alpha}=(0.975, 0.975, 3.0)$ (right)}
    \label{fig:dirichlet_illustration}
\end{figure}

Figure \ref{fig:dirichlet_illustration} presents examples of Dirichlet distributions on the unit simplex of $\R^3$.

\subsection{Majorization-Minimization algorithm}\label{app:majorant_F}


We provide the details for our new MM Algorithm \ref{algo:MM_update_alpha} for minimizing \eqref{eq:def_negative_log_likelihood_k}. Our approach is based on constructing a quadratic bound of the function $\ln \Gamma( \cdot +1)$, which is a consequence of the following lemma. 

\begin{lemma}[\cite{erdogan2002monotonic}]\label{lemma:specific_quad_maj}
    Let $\psi$ be a twice-continuously differentiable function on $[0,+\infty[$.
    Assume that $\psi''$ is decreasing on $[0,+\infty[$. Let $z\in [0,+\infty[$
    and let 
    \begin{equation}
        c_\psi(z) = \begin{cases}
        \psi''(0) & \mbox{if $z = 0$}\\
        \displaystyle 2\frac{\psi(0)-\psi(z)+\psi'(z)z}{z^2} & \mbox{otherwise.}
        \end{cases}
    \end{equation}
    Then, for every $x\in [0,+\infty[$,
    \begin{equation}
    \psi(x) \le \psi(z) + \psi'(z)(x-z)
    + \frac12 c_\psi(z) (x-z)^2.
    \end{equation}
\end{lemma}

We are now ready to prove Lemma \ref{lemma:majorant_F}.


\begin{proof}
We first observe that $\bs{\alpha}_k \mapsto -\ln \Gamma \left(\sum_{i=1}^K \alpha_{k,i} \right)$ is concave. Consequently, we can upper-bound this term at $\bs{\beta}_k$ using its first-order Taylor expansion around $\bs{\beta}_k$. Furthermore, considering the relation
\begin{equation}
\forall t \in (0, +\infty), \quad \ln \Gamma(t) = \varphi(t) - \ln t,
\end{equation}
and given that the prerequisites of Lemma \ref{lemma:specific_quad_maj} are fulfilled by $\varphi$, the result in \eqref{eq:majorant_F} follows immediately.
\end{proof}

For a fixed value of $\bs{\beta}_k \in (0, +\infty)^K$, the minimizer $\widehat{\bs{\alpha}}_k$ of the majorant given by Lemma \ref{lemma:majorant_F} is such that, for every $i\in \{1, \dots, K\}$, $\widehat{\alpha}_{k,i}$ is the unique positive root of the second order polynomial equation
\begin{equation}
c(\beta_{k,i}) \alpha_{k,i}^2
+ b_{k,i}(\boldsymbol{\beta}_{k})
\alpha_{k,i} = 1,
 \end{equation}
with
\begin{multline}
b_{k,i}(\bs{\beta}_k)  = \varphi'(\beta_{k,i})
- (\ln \Gamma)'\Big(\sum_{j=1}^K \beta_{k,j}\Big)
-c(\beta_{k,i}) \beta_{k,i}\\
- \left(\sum_{n=1}^N u_{n,k}\right)^{-1}\sum_{n=1}^N u_{n,k} \ln z_{n,i}.
\end{multline}
Hence,
\begin{equation}\label{eq:update_MM_alpha}
\widehat{\alpha}_{k,i}
= \frac{-b_{k,i}(\bs{\beta}_k)+
\sqrt{\big(b_{k,i}(\bs{\beta}_k)\big)^2
+ 4 c(\beta_{k,i})}}
{2c(\beta_{k,i})},
\end{equation}
which yields the MM updates described in Algorithm \ref{algo:MM_update_alpha}.

In Table \ref{table:comparison_time}, we compare the convergence speed of the MM Algorithm \ref{algo:MM_update_alpha} and the Block MM Algorithm \ref{algo:em_dirichlet}, using our majorant \eqref{eq:majorant_F} versus the one proposed by Minka in \cite{minka2000estimating}. For Algorithm \ref{algo:MM_update_alpha}, the convergence criterion is defined as $\frac{\Vert \boldsymbol{\alpha}^{(m+1)} - \boldsymbol{\alpha}^{(m)} \Vert^2}{\Vert \boldsymbol{\alpha}^{(m)} \Vert^2} \leq \varepsilon$, and for Algorithm \ref{algo:em_dirichlet} as $\frac{\Vert \boldsymbol{\alpha}^{(\ell+1)} - \boldsymbol{\alpha}^{(\ell)} \Vert^2}{\Vert \boldsymbol{\alpha}^{(\ell)} \Vert^2} \leq \varepsilon$, where $\varepsilon = 10^{-13}$. Our MM algorithm is approximately twice as fast as Minka's.

\begin{table}[h]
    \centering
    \begin{tabular}{l|cc}
        & Algo. \ref{algo:MM_update_alpha} & Algo. \ref{algo:em_dirichlet}\\ \hline
       Minka's \cite{minka2000estimating}  & $2.04\times 10^{-1}$ & $2.09$\\
      \rowcolor{lightsalmon!20} Ours  & $7.62 \times 10^{-2}$ & $1.04$
    \end{tabular}
    \caption{Time before reaching the convergence criterion in seconds, for Algorithm \ref{algo:MM_update_alpha} and \ref{algo:em_dirichlet}. The displayed time is the average execution time per task, computed over 1,000 tasks, on the ImageNet dataset with 4 shots. }
    \label{table:comparison_time}
\end{table}


\subsection{Estimation step on assignments in our algorithm}\label{app:details_step_u}

We provide more details on the derivation of the closed-form update of variable $\boldsymbol{u}_n$ at each iteration $\ell \in \N$. Consider the function $F$ given by
\begin{multline}\label{eq:F_update_u}
 F(\boldsymbol{u}_n) = -\sum_{k=1}^K u_{n,k} \ln \left( \p \left( \boldsymbol{z}_{n} \mid \boldsymbol{\alpha}_k \right) \right) + \iota_{\Delta_K}(\boldsymbol{u}_n) \\
 -\frac{\lambda}{|\query|} (\ln(\boldsymbol{\pi}^{(\ell+1)}) + \boldsymbol{1})^\top (\boldsymbol{u}_n - \boldsymbol{u}_n^{(\ell)}) + \sum_{k=1}^K u_{n,k} \ln u_{n,k},
\end{multline}
where $\iota_{\Delta_K}$ is the indicator function of the simplex $\Delta_K$, assigning zero to points within the simplex and $+\infty$ elsewhere. 

Let us see how to compute the minimizer of \eqref{eq:F_update_u} via the proximal operator (see \cite[Eq. 24.2]{bauschke2019convex} for a definition). We define the function $\psi$ on $\mathbb{R}^K$ as
\begin{equation}
   \psi(\boldsymbol{x}) = \left\{ \begin{array}{ll}  
   \sum_{k=1}^K x_k \ln(x_k) - \frac{x_k^2}{2}, & \text{if } \boldsymbol{x} \in \Delta_K, \\
   +\infty, & \text{otherwise.} 
   \end{array} \right.
\end{equation}
The proximal operator of $\psi$, which is well-established as the softmax function, allows for the practical computation of the minimizer \cite[Example 2.23]{combettes2020deep}.
Since $F$ is proper, lower semi continuous and convex, finding the minimizer of $F$ is equivalent to finding $\bs{u}_n$ such that $0 \in \partial F(\bs{u}_n)$. This reads

\begin{align}
    & 0 \in \partial F(\bs{u}_n)\nonumber\\
\iff \quad & 0 \in -  \ln( \p \left( \boldsymbol{z}_{n} ~|~ {\boldsymbol{\alpha}_k} \right )) -\frac{\lambda}{|\query|} (\ln(\bs{\pi}^{(\ell+1)})+1) \nonumber \\ & \qquad  +\partial \psi(\bs{u}_n) + \bs{u}_n \nonumber ,\\
\iff \quad & \bs{u}_n =  \text{\normalfont softmax}\left(  \left(\ln \p \left( \boldsymbol{z}_{n} ~|~ {\boldsymbol{\alpha}_k}  \right)+ \frac{\lambda}{|\query|} \ln \pi_{k}^{(\ell+1)} \right)_{k}  \right),\nonumber
\end{align}
where we used the characterization of the proximity operator \cite[Prop. 16.44]{bauschke2019convex}.

\subsection{Class-assignment in the zero-shot setting}\label{app:class_assignment}

\begin{figure}[h!]
    \centering
    \includegraphics[scale=0.7]{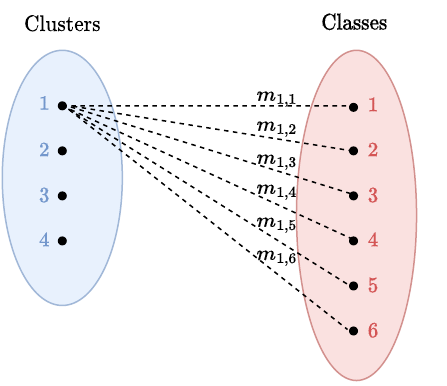}
    \caption{Illustration of the bipartite matching for class assignment. \label{fig:bipartite_matching}}
    \label{fig:enter-label}
\end{figure}

Figure \ref{fig:bipartite_matching} gives an illustration of our graph matching procedure for assigning each cluster to a unique class. 

Note that it is possible to not perform the graph matching procedure and simply assign to each cluster $k \in \mathcal{K}$ the class $\ell^* \in \{1, \dots, K\}$ such that $\ell^* = \underset{\ell \in \{1, \dots, K\}}{\argmax} \, m_{k,\ell}$, where $\mathbf{m}_k = (m_{k,\ell})_{1 \leq \ell \leq K}$ is the average of simplex features assigned to cluster $k$. However, this leads in practice to multiple clusters being assigned to the same class. We nevertheless provide the zero-shot accuracy results in Table \ref{table:zero-shot_without_graph}.

\subsection{Links with the EM algorithm}\label{app:links_EM}

We give a proof of Proposition \ref{prop:links_EM}.

\begin{proof}
Given the mixture model \eqref{eq:dirichlet_mixture}, the EM algorithm aims at maximizing the log-likelihood function 
\begin{equation}
    L(\bs{\pi}, \bs{\alpha}) =  \sum_{n \in \query} \ln\left( \sum_{k=1}^K \pi_k \p \left( \boldsymbol{z}_{n} ~|~ {\boldsymbol{\alpha}_k} \right )  \right)
\end{equation}
with respect to $\bs{\pi}$ and $\bs{\alpha}$. The process involves two steps: \textbf{expectation} and \textbf{maximization}, and the algorithm iteratively generates sequenc\textbf{}es $\{\bs{\pi}^{(\ell)}\}_{\ell \in \N} \subset \Delta_K$ and, for every $k \in \{1, \dots, K\}$, $\{\bs{\alpha}_k^{(\ell)}\}_{\ell \in \N} \subset (0, +\infty)^K$.

During the \textbf{expectation step}, for a given iteration number $\ell \in \N$, we compute the expected responsibilities. For each query sample $n \in \query$, we define $\bs{u}_n^{(\ell)} = (u_{n, k}^{(\ell)})_{ 1 \leq k \leq K} $ by
\begin{equation}\label{eq:def_u_tight}
    u_{n, k}^{(\ell)} =  \frac{\pi_k^{(\ell)} \p \left( \boldsymbol{z}_{n} ~|~ {\boldsymbol{\alpha}_k^{(\ell)} } \right ) }{\sum_{i=1}^K \pi_i^{(\ell)} \p \left( \boldsymbol{z}_{n} ~|~ {\boldsymbol{\alpha}_i^{(\ell)} } \right ) }.
\end{equation}
This quantity corresponds to the probability of the data point $n$ belonging to class $k$ based on the current estimates of $\bs{\pi}^{(\ell)}$ and $\bs{\alpha}_k^{(\ell)}$.

In the \textbf{maximization step}, we derive an upper bound for the log-likelihood at the current iterate using the responsibilities calculated in the expectation step, along with Jensen's inequality. This majorization reads 
\begin{equation}
    L(\bs{\pi}, \bs{\alpha}) \leq q((\bs{\pi}, \bs{\alpha}); (\bs{\pi}^{(\ell)}, \bs{\alpha}^{(\ell)}) ),
\end{equation}
where $q(\,\cdot\,; (\bs{\pi}^{(\ell)}, \bs{\alpha}^{(\ell)}) )$ is defined, for all $\bs{\pi}\in \Delta_K$ and $\bs{\alpha}\in ((0, +\infty)^K)^K$, by
\begin{equation*}\label{eq:def_surrogate}
   q((\bs{\pi}, \bs{\alpha}); (\bs{\pi}^{(\ell)}, \bs{\alpha}^{(\ell)}) ) = \sum_{n \in \query} \sum_{k=1}^K u_{n,k}^{(\ell)}\ln \left(\frac{\pi_k \p \left( \boldsymbol{z}_{n} ~|~ {\boldsymbol{\alpha}_k}  \right )}{u_{n, k}^{(\ell)}}  \right).
\end{equation*}
This upper bound is separable and defines a tight majorant, i.e., $q((\bs{\pi}^{(\ell)}, \bs{\alpha}^{(\ell)}); (\bs{\pi}^{(\ell)}, \bs{\alpha}^{(\ell)}) ) =L(\bs{\pi}^{(\ell)}, \bs{\alpha}^{(\ell)})$. 
Next, one maximizes the majorant with respect to $\bs{\alpha}$ and $\bs{\pi}$ under the simplex constraints. This yields the expression
\begin{equation}
\label{eq:update-label-marginal}
(\forall k \in \{1, \dots, K\})\quad {\pi}^{(\ell+1)}_{k} = \frac{1}{|\query|} \sum_{n \in \query} u_{n,k}^{(\ell)},
\end{equation}
i.e., the mixing coefficients are the average of the responsibilities for each class over all data points in the query set. On the other hand, for each class $k\in\{1, \dots, K\}$, the parameters $\bs{\alpha}_k^{(\ell+1)}$ are set by solving the optimization problem
\begin{equation}
    \underset{\bs{\alpha}_k \in (0, +\infty)^K}{\mathrm{maximize}} \,  \sum_{n \in \query} u_{n,k}^{(\ell)}\ln \p \left( \boldsymbol{z}_{n} ~|~ {\boldsymbol{\alpha}_k}  \right ).
\end{equation}

We can then show that the updates are identical to those performed in Algorithm \ref{algo:em_dirichlet} when $\lambda = |\query|$ and $\support = \varnothing$. The identity of the updates on $\bs{\alpha}$ and $\bs{\pi}$
are obvious.
For $\bs{u}$, note that Equation \eqref{eq:def_u_tight} can be rewritten 
\begin{align}
    u_{n, k}^{(\ell+1)} &= \frac{\pi_k^{(\ell+1)} \p ( \boldsymbol{z}_{n} ~|~ {\boldsymbol{\alpha}_k^{(\ell+1)} } ) }{\sum_{i=1}^K \pi_i^{(\ell+1)} \p ( \boldsymbol{z}_{n} ~|~ {\boldsymbol{\alpha}_i^{(\ell+1)} } ) },\nonumber\\
    &= \frac{\exp\left(\ln \pi_k^{(\ell+1)} + \ln \p ( \boldsymbol{z}_{n} ~|~ {\boldsymbol{\alpha}_k^{(\ell+1)} }) \right)}{\sum_{i=1}^K \exp\left( \ln \pi_i^{(\ell+1)} + \ln \p ( \boldsymbol{z}_{n} ~|~ {\boldsymbol{\alpha}_i^{(\ell+1)} } )\right) },\nonumber
\end{align}
or equivalently, 
\begin{equation}
    \bs{u}_n = \mathrm{softmax}\left((\ln \pi_k^{(\ell+1)} + \ln \p ( \boldsymbol{z}_{n} ~|~ \boldsymbol{\alpha}_k^{(\ell+1)}))_k \right),
\end{equation}
thus aligning with the update in Algorithm \ref{algo:em_dirichlet}.

\end{proof}

\subsection{Zero-shot performance as a function of the size the query set}\label{app:variation_query_size}

\begin{figure*}[h]
    \centering
    \includegraphics[scale=0.5]{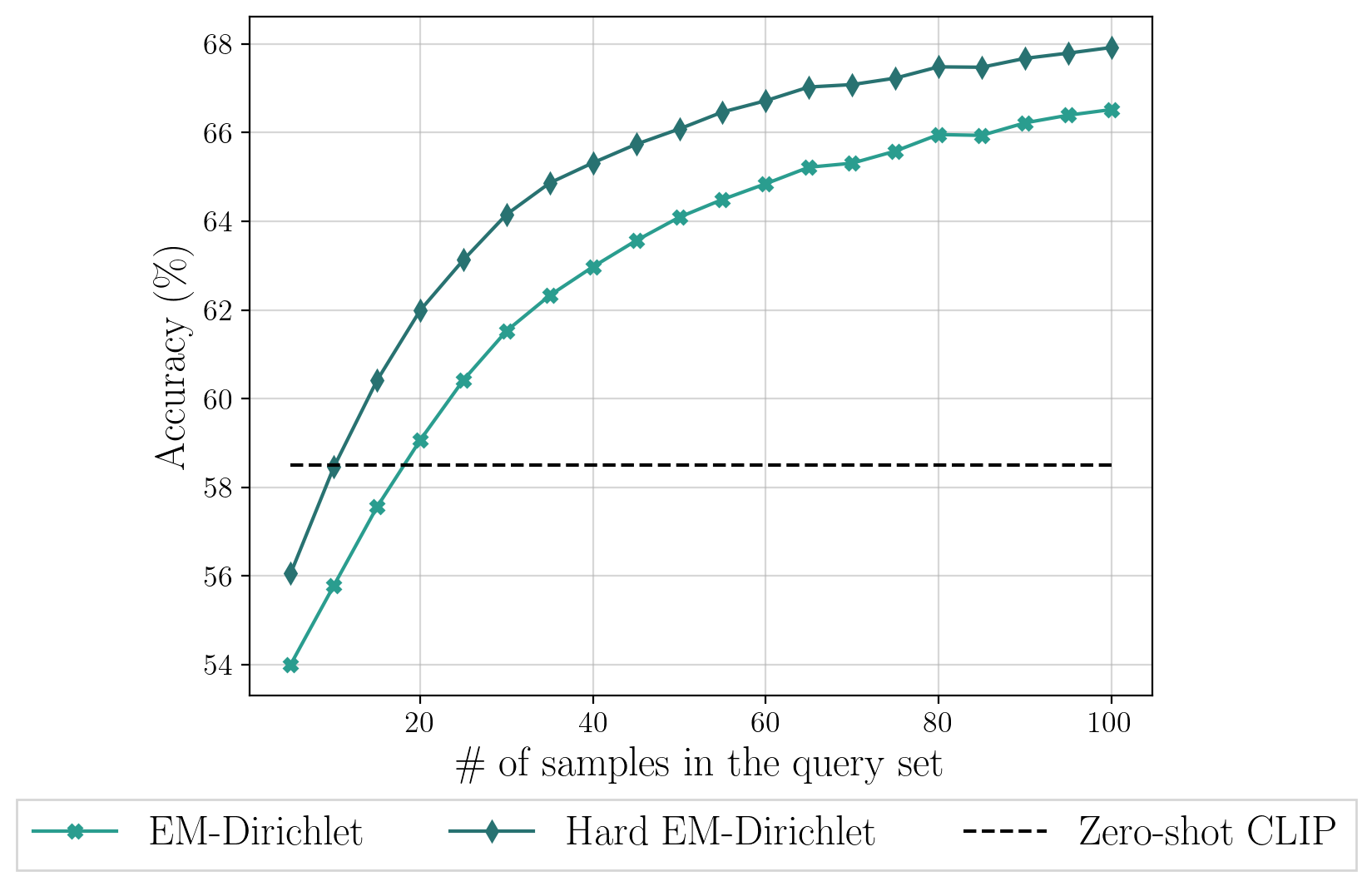}
    \caption{Average accuracy on the 11 datasets as a function of the number of samples in the query set, over 1,000 tasks generated following the protocol described in Section \ref{sec:zero_shot_generation}. As anticipated, the efficiency of transduction increases with the number of samples in the query set.}
    \label{fig:accuracy_vs_query}
\end{figure*}

We point to Figure \ref{fig:accuracy_vs_query} which displays the accuracy of our methods EM-Dirichlet and Hard EM-Dirichlet in the zero-shot setting versus the number of samples in the query set.

\subsection{Additional results in the few-shot setting}\label{app:additional_results_few_shot}

\begin{figure*}[ht]
    \centering
    \includegraphics[scale=0.6]{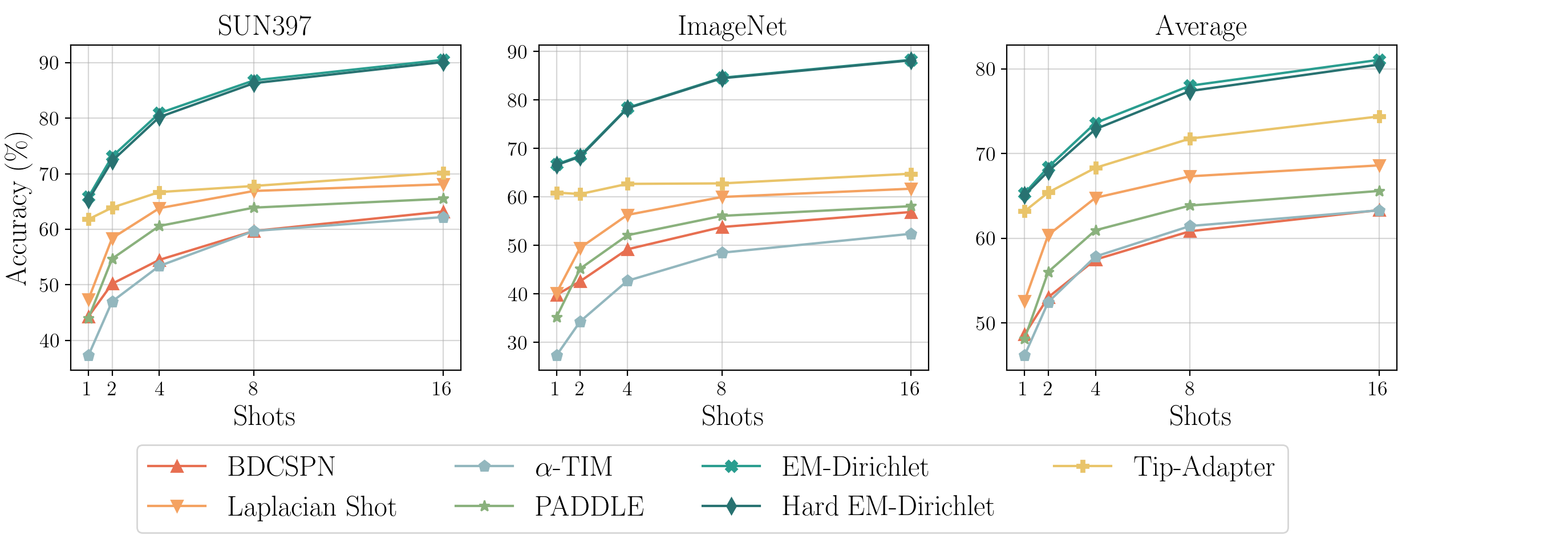}
    \caption{Accuracy versus shots for seven methods from Table \ref{table:few-shot} on SUN397, ImageNet, and the average across the 11 datasets.}
    \label{fig:few-shot_average}
\end{figure*}

\begin{figure*}[ht]
    \centering
    \includegraphics[scale=0.6]{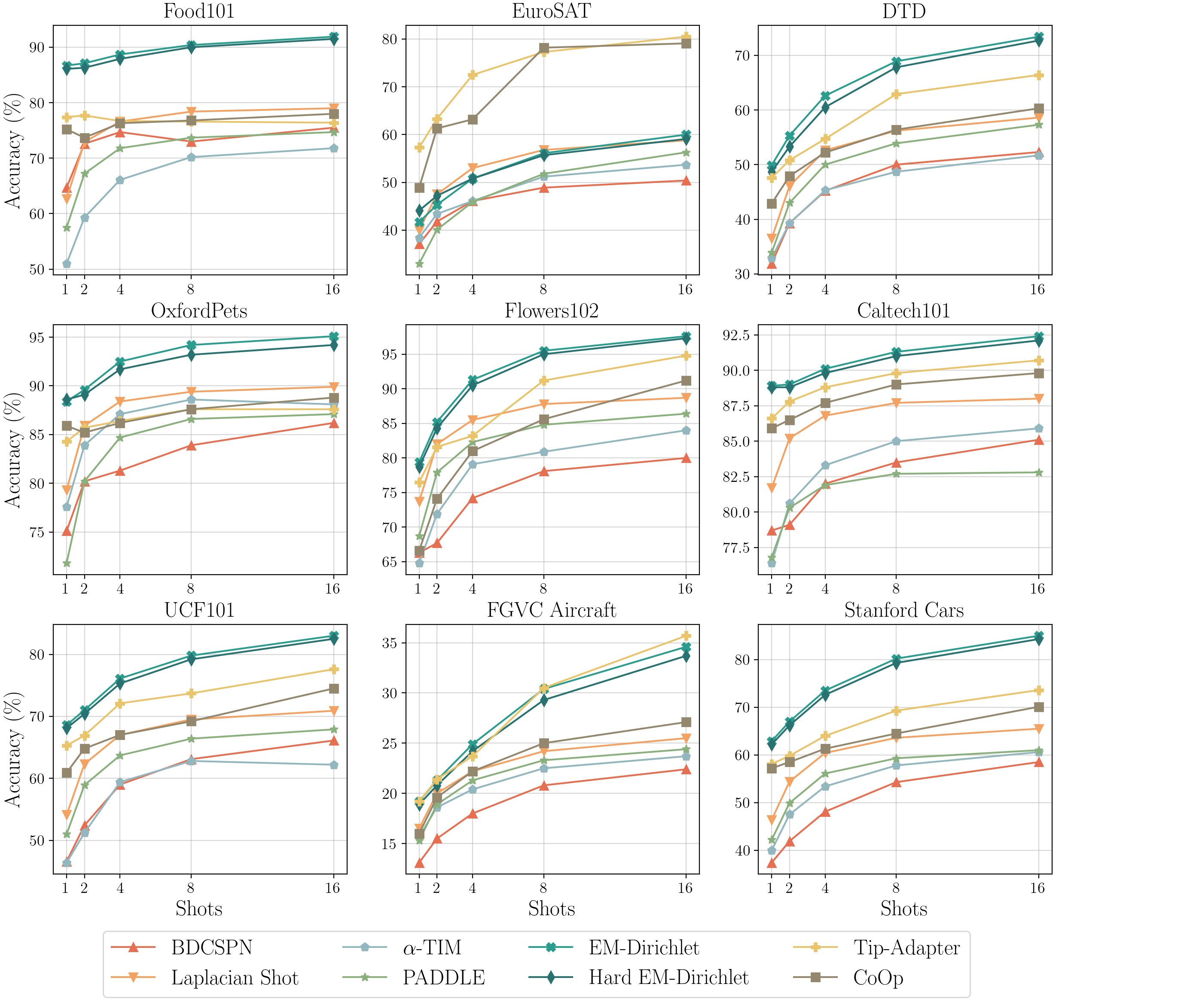}
    \caption{Accuracy versus shots for eight methods from Table \ref{table:few-shot} on 9 datasets.}
    \label{fig:few-shot_all}
\end{figure*}

In addition to the results in the 4-shot case presented in Table \ref{table:few-shot}, we provide the results for other number of shots. Figure \ref{fig:few-shot_average} displays the accuracy as a function of the number of shots. This analysis includes our methods EM-Dirichlet and Hard EM-Dirichlet, other transductive methods (BDCSPN, Laplacian Shot, $\alpha$-TIM, PADDLE), and the inductive Tip-Adapter method. We did not evaluate CoOp because of the prohibitive time required to run the method, as underlined in Table \ref{table:few-shot}. We observe that our method significantly outperforms its closest competitor, TIP, on the challenging SUN397 and ImageNet datasets, as well as on the average of the 11 datasets. This gap gets even wider when the number of shots increases. Complete results for all datasets are given in Figure \ref{fig:few-shot_all}.

\subsection{Ablation study on each term of the objective}

We provide an ablation study on our objective function, which minimizes $-\mathcal{L} + \Phi + \Psi$ under simplex constraints, where $\mathcal{L}$ is the log-likelihood, $\Phi$ a barrier term, and $\Psi$ a partition complexity term promoting fewer clusters. 
Note that, when removing barrier term $\Phi$, our update step for the assignment variables (Eq. \eqref{eq:cost_function_wrt_u} without the barrier term) amounts to solving a linear programming problem, resulting in integer solutions (i.e., hard assignments), akin to what we coined ``Hard EM-Dirichlet''.

Table \ref{table:ablation} demonstrates the effect of each term. The partition complexity term $\Psi$ significantly enhances performance. In contrast, the barrier term $\Phi$, in isolation, does not improve performance. However, when combined with $\Psi$, it shows utility in the 4-shot scenario. The inclusion of $\Phi$ was primarily to maintain a soft assignment approach and to make the link with the EM algorithm (Proposition \ref{prop:links_EM}).

\begin{table}[htb]
  \centering
  \resizebox{0.7\hsize}{!}{
    \begin{tabular}{ll|cc|c}
                       & \textbf{Criterion} &{\textbf{Acc.}} \\ \hline
    \multirow{4}{*}{\rotatebox[origin=c]{90}{\textcolor{blue}{0-shot}}} & {$-\mathcal{L}$}    &  50.8\\
    & {$-\mathcal{L} + \Phi$}                                                                   &  42.7\\
    & {$-\mathcal{L} + \Psi \quad$ (= Hard EM-Dirichlet)}                                                     &  67.6\\
    & {$-\mathcal{L} + \Phi + \Psi \quad$ (= EM-Dirichlet)}                                                     & 65.8 \\ \hline 
    \multirow{4}{*}{\rotatebox[origin=c]{90}{\textcolor{blue}{4-shot}}} & {$-\mathcal{L}$}       & 59.5\\
    & {$-\mathcal{L} + \Phi$}                                                                  & 58.8 \\
    & {$-\mathcal{L} + \Psi \quad$ (= Hard EM-Dirichlet)}                                                    & 72.9\\
    & {$-\mathcal{L} + \Phi + \Psi \quad$ (= EM-Dirichlet)}                                                     & 73.6\\
    \end{tabular}}
    \caption{Average accuracy on the 11 datasets, over 1,000 classification tasks. Inference is performed on the text-vision probability features.}
    \label{table:ablation}
\end{table}

\subsection{Using the similarity scores as feature vectors}

One might consider directly using the visual-textual embeddings as input features (specifically, the cosine similarities) without applying a softmax function. It could be hypothesized that methods targeting a Gaussian distribution might perform more effectively with these raw features than with probability features. However, as indicated in Table \ref{table:without_softmax}, this is not the case. Employing a Gaussian distribution within the joint visual-textual embedding space actually leads to decreased accuracy when compared to our method that utilizes probability features.

\begin{table}[htb]
  \centering
  \resizebox{0.7\hsize}{!}{
    \begin{tabular}{l|c|c}
  \textbf{Method} & \textbf{Acc.} & \textbf{Loss in acc.}\\ \hline
  Soft K-means & 28.2 & 2.1\\
EM-Gaussian (diag. cov.) & 34.9 & 14.8
    \end{tabular}}
    \caption{Average accuracy on the 11 datasets, over 1,000 zero-shot tasks using text-vision features (without softmax). The accuracy loss is measured against the results with probability features. \label{table:without_softmax}}
\end{table}

\begin{table*}
  \centering
    \begin{tabular}{ll|ccccccccccc|c}
                       &  & \rotatebox[origin=l]{90}{Food101} & \rotatebox[origin=l]{90}{EuroSAT} &  \rotatebox[origin=l]{90}{DTD} &\rotatebox[origin=l]{90}{OxfordPets}  & \rotatebox[origin=l]{90}{Flowers102}  & \rotatebox[origin=l]{90}{Caltech101}  & \rotatebox[origin=l]{90}{UCF101}  & \rotatebox[origin=l]{90}{FGVC Aircraft} &\rotatebox[origin=l]{90}{Stanford Cars} &\rotatebox[origin=l]{90}{SUN397} &\rotatebox[origin=l]{90}{ImageNet} &\rotatebox[origin=l]{90}{\textbf{Average}} \\ \hline
    &     Zero-shot CLIP & 77.1 & 36.5 & 42.9 & 85.1 & 66.1 & 84.4 & 61.7 & 17.1 & 55.8 & 58.6 & 58.3 & 58.5\\ \hline
    \multirow{4}{*}{\rotatebox[origin=c]{90}{\textcolor{blue}{Vis. embs.}}} & {Hard K-means}    & 78.4 & 34.5 & 46.2 & 86.3 & 70.2 & 87.3 & 66.1 & 19.2 & 58.7 & 62.9 & 60.9 & 61.0\\
    & {Soft K-means}                                                                 & 79.3 & 28.4 & 42.8 & 67.5 & 64.7 & 86.0 & 62.7 & 17.7 & 57.5 & 59.0 & 59.3 & 56.8\\
    & {EM-Gaussian ($\mathrm{Id}$ cov.)}                                                   & 14.0 & 14.5 & 9.4 & 6.9 & 5.3 & 30.3 & 7.4 & 1.9 & 2.5 & 5.3 & 3.9 & 8.3\\   
    & {EM-Gaussian (diag cov.)}                                                   & 77.1 & \textbf{37.1} & 44.1 & 86.9 & 68.9 & 85.8 & 63.8 & 18.4 & 57.3 & 60.1 & 59.3 & 59.9\\ \hline 

    \multirow{6}{*}{\rotatebox[origin=l]{90}{\textcolor{blue}{Probabilities}}} & {Hard K-means}       & 80.2 & 34.7 & 45.9 & 88.7 & 69.0 & 86.9 & 66.6 & 20.1 & 59.7 & 63.7 & 61.0 & 61.5\\
    & {Soft K-means}                                                                 & 43.4 & 22.1 & 18.7 & 67.7 & 36.2 & 54.7 & 31.7 & 7.6 & 36.3 & 18.9 & 19.1 & 32.4\\
     & {EM-Gaussian ($\mathrm{Id}$ cov.)}                                                                & 21.4 & 14.5 & 16.5 & 21.1 & 23.1 & 33.6 & 19.3 & 6.8 & 18.5 & 18.7 & 19.1 & 19.3\\ 
     & {EM-Gaussian (diag cov.)}                                                   & 78.9 & 33.4 & 44.8 & 87.9 & 69.3 & 86.6 & 65.7 & 20.2 & 63.5 & 66.1 & 63.0 & 61.8\\ 
    & {Hard KL K-means}                                                                   & 84.3 & 34.4 & 46.2 & 90.3 & 72.3 & 88.3 & 69.5 & 21.4 & 68.6 & 62.4 & 61.0 & 63.5\\
        \rowcolor{lightsalmon!20}  & EM-Dirichlet                                & 89.0 & 32.9 & 48.7 & 91.2 & 73.1 & 90.4 & 70.5 & 21.4 & 69.5 & 78.1 & 78.0 & 67.5\\
       \rowcolor{lightsalmon!40}        &    Hard EM-Dirichlet                   & \textbf{90.7} & 33.5 & \textbf{49.8} & \textbf{92.6} & \textbf{73.9} & \textbf{91.1} & \textbf{71.3} & \textbf{22.0} & \textbf{70.8} & \textbf{79.1} & \textbf{78.5} & \textbf{68.5}\\
    \end{tabular}
    \caption{Evaluation of the methods computing the accuracy without the graph matching. Average accuracy of clustering methods over 1,000 zero-shot classification tasks. Inference is performed both on the visual embeddings and on the text-vision probability features.}
    \label{table:zero-shot_without_graph}
\end{table*}

\end{document}